\documentclass[10pt,twocolumn,letterpaper]{article}

\usepackage[pagenumbers]{wacv} 

\usepackage{graphicx}
\usepackage{amsmath}
\usepackage{amssymb}
\usepackage{nicematrix}
\usepackage{booktabs}
\usepackage{threeparttable}  

\usepackage[pagebackref,breaklinks,colorlinks]{hyperref}
\usepackage{mathtools}
\usepackage{graphicx}
\usepackage{algpseudocode}
\usepackage{caption}
\usepackage{subcaption}
\usepackage{multirow}
\usepackage{tablefootnote}
\usepackage{wrapfig}
\usepackage{textcomp}
\usepackage{siunitx}
\usepackage{multirow}
\usepackage{wrapfig}
\usepackage{booktabs}
\usepackage{graphics}
\usepackage{multibib}
\usepackage{algorithm}
\usepackage{arydshln}

\newcommand{\opneuromodulation}{\times^{\odot}}

\usepackage[capitalize]{cleveref}
\crefname{section}{Sec.}{Secs.}
\Crefname{section}{Section}{Sections}
\Crefname{table}{Table}{Tables}
\crefname{table}{Tab.}{Tabs.}

\def\wacvPaperID{2215} 
\def\confName{WACV}
\def\confYear{2024}

\begin{document}

\title{Learning to Modulate Random Weights: Neuromodulation-inspired Neural Networks For Efficient Continual Learning}

\author{
Jinyung Hong\textsuperscript{1} and Theodore P.~Pavlic\textsuperscript{1, 2, 3, 4} \\
\textsuperscript{1} School of Computing and Augmented Intelligence\\
\textsuperscript{2} School of Sustainability \\
\textsuperscript{3} School of Complex Adaptive Systems \\
\textsuperscript{4} School of Life Sciences\\
Arizona State University\\
Tempe, AZ 85281\\
{\tt\small \{ jhong53, tpavlic \}@asu.edu}
}
\maketitle

\begin{abstract}
Existing Continual Learning~(CL) approaches have focused on addressing catastrophic forgetting by leveraging regularization methods, replay buffers, and task-specific components. However, realistic CL solutions must be shaped not only by metrics of catastrophic forgetting but also by computational efficiency and running time. Here, we introduce a novel neural network architecture inspired by neuromodulation in biological nervous systems to economically and efficiently address catastrophic forgetting and provide new avenues for interpreting learned representations. \emph{Neuromodulation} is a biological mechanism that has received limited attention in machine learning; it dynamically controls and fine tunes synaptic dynamics in real time to track the demands of different behavioral contexts. Inspired by this, our proposed architecture learns a relatively small set of parameters per task context that \emph{neuromodulates} the activity of unchanging, randomized weights that transform the input. We show that this approach has strong learning performance per task despite the very small number of learnable parameters. Furthermore, because context vectors are so compact, multiple networks can be stored concurrently with no interference and little spatial footprint, thus completely eliminating catastrophic forgetting and accelerating the training process. 
All codes are available at~\url{https://github.com/jyhong0304/CRWN_CL}.
\end{abstract}

\begin{figure*}[t!]\centering%
\begin{subfigure}[t]{0.47\textwidth}\centering
\includegraphics[width=\textwidth]{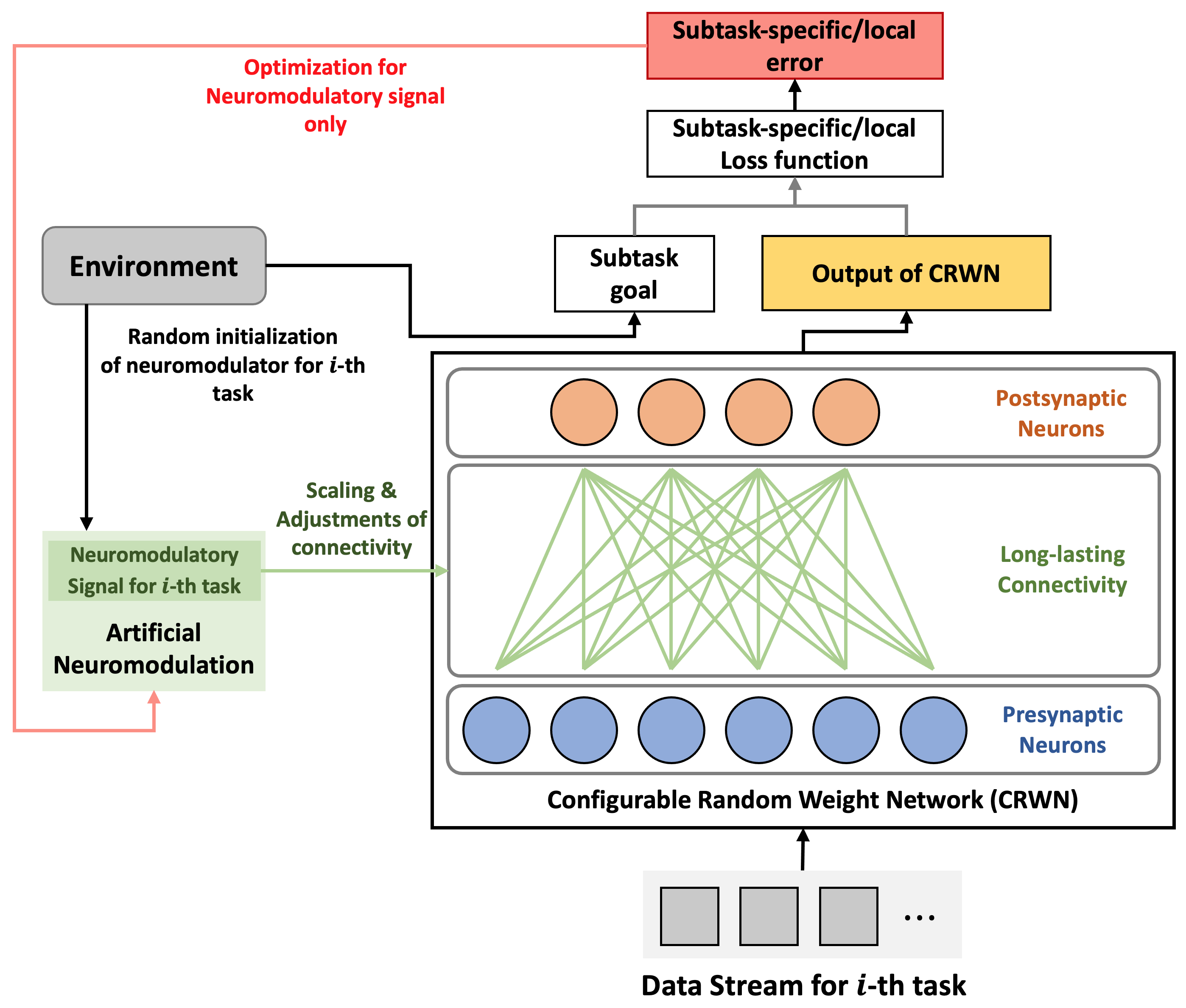}
\caption{The proposed conceptual framework of neuromodulatory process for continual learning, inspired by~\protect\cite[Fig.~3]{mei2022informing}.}
\label{fig:neuromodulation_framework}
\end{subfigure}
\hfill
\begin{subfigure}[t]{0.45\textwidth}\centering
\includegraphics[width=\textwidth]{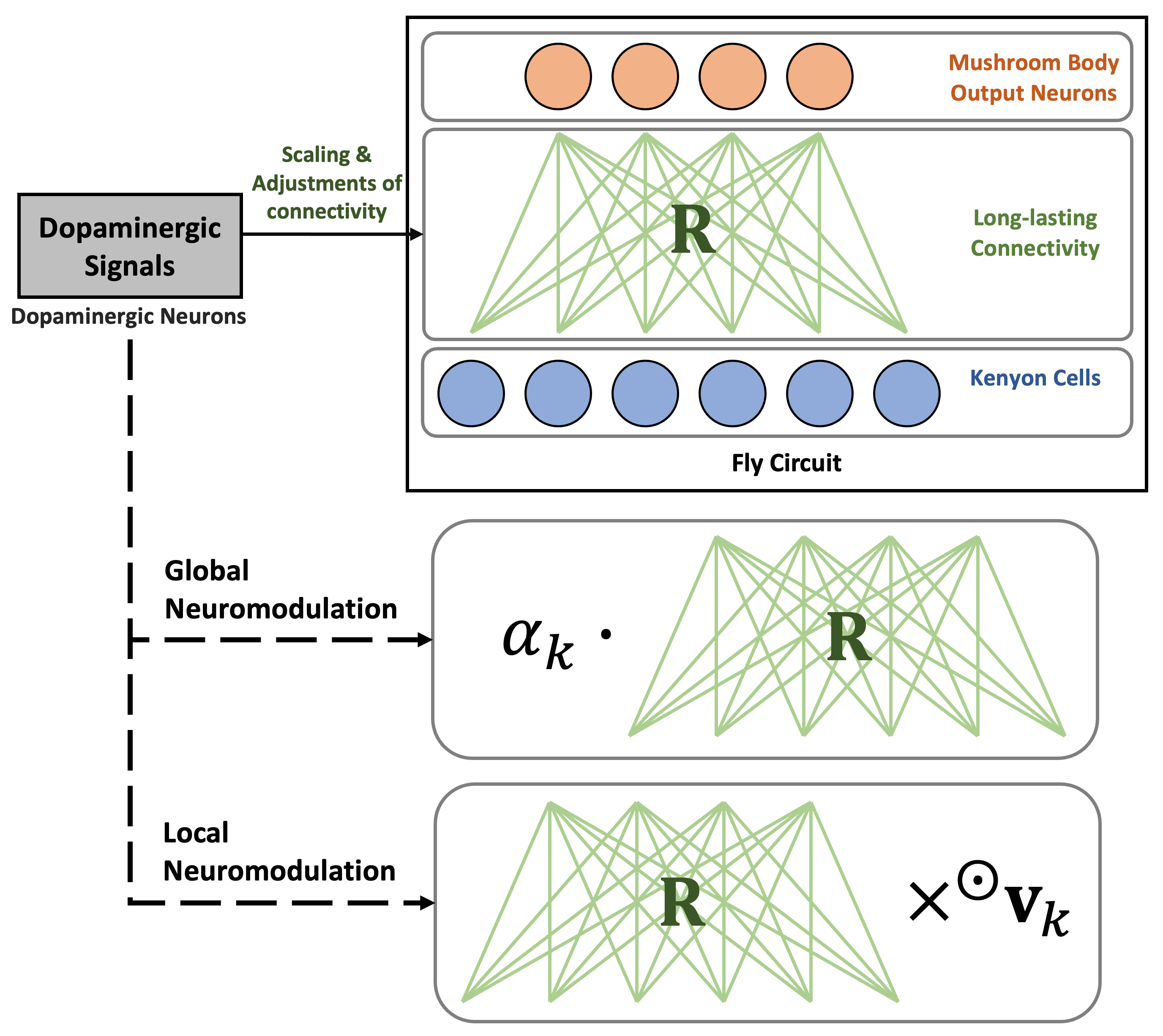}
\caption{Action of dopamine and memory module motif of the fly circuit~(\textbf{top}) and two types of neuromodulation~(\textbf{bottom}). The goal of our approach is to learn to modulate fixed, random weights $\textbf{R}$ for the $k$-th task, which allows for optimizing the scalar $\alpha_{k}$~(\textbf{Global}) and the vector $\textbf{v}_{k}$~(\textbf{Local}) only. See Eq.~(\ref{eq:neuromodulation}).}
\label{fig:neuromodulation_types}
\end{subfigure}
\hfill
\caption{\emph{The overall architecture of the proposed artificial neuromodulation for continual learning}}
\label{fig:neuromodulation}
\end{figure*}
\section{Introduction}
\label{sec:introduction}
The main goal of systems for Continual Learning~(CL), or Lifelong Learning~\cite{thrun1995lifelong}, is to learn a series of tasks sequentially, mimicking the human ability to learn new concepts over a lifetime in an incremental way. Thus, an ideal continual learner will utilize the learned knowledge from previous tasks when solving new ones and update previous task knowledge with information learned in new tasks. However, continual learning in artificial neural networks have historically been prone to \emph{catastrophic forgetting} or \emph{catastrophic interference}, where the model performance on previous tasks significantly decreases upon learning new tasks~\cite{mcclelland1995there}. The various approaches that have been proposed to address catastrophic forgetting can generally be grouped into the three categories of: 
i) \textbf{regularization-based methods} that aim to retain the learned information of past tasks during continual learning with the help of elaborately designed regularization terms~\cite{chaudhry2020continual, jung2020continual, mirzadeh2020linear}; ii) \textbf{rehearsal/replay-buffer-based methods} that utilize and revisit a set of actual or synthesized samples from the previous tasks~\cite{chaudhry2019continual, chaudhry2019tiny, saha2021gradient}; and iii) \textbf{architecture/task-specific-components-based methods} that minimize the interference between tasks through specially designed architectural components~\cite{cheung2019superposition, wen2020batchensemble, wortsman2020supermasks, kang2022forget}.

Despite the success of recent work on reducing catastrophic forgetting, these methods still lack many  properties desirable in CL systems, such as high computational efficiency and fast execution time. For example, because regularization-based approaches do not ensure compressed task representations, they inevitably forget or stop learning due to structural rigidity and fixed capacity as the size and variety of learning tasks increase. Also, rehearsal-based methods share the same drawback as regularization methods and require additional storage to store the replay buffer or generative models. On the other hand, architecture-based methods can mitigate limited capacity directly by enabling the architecture to grow over time. However, finding efficient ways of utilizing task-specific components can be another burden. Some recent studies of CL methods~\cite{wortsman2020supermasks, kang2022forget} based on the Lottery Ticket Hypothesis~(LTH)~\cite{frankle2018lottery} have shown the effectiveness of searching for the optimal subnetworks during CL. However, searching for the subnetwork with iterative pruning needs to optimize the weights of the exact size as the learnable parameters of the original dense network, which is impractical and inefficient.

We depart from the above perspectives and focus on a completely different way to address catastrophic forgetting inspired by the neuromodulatory system in the biological nervous system. \emph{Neuromodulators}, including acetylcholine~(ACh), noradrenaline~(NA), serotonin~(5-HT), and dopamine~(DA), are chemical messengers that diffusely target neurons and synapses in specific brain regions, found not only in humans~\cite{mei2022informing} but also in the brains of insects such as fruit flies~\cite{modi2020drosophila}. They play a critical role in the organization in the neuromodulatory systems across different temporal and spatial scales, but we focus on the neuromodulatory process that directly influences the connection between presynaptic and postsynaptic neurons~(Fig.~\ref{fig:neuromodulation_framework}) providing high-level contextual control of cognition and behavior~\cite{mei2022informing}. This contextual control operates on the one hand by reshaping local cellular and synaptic physiology and, on the other hand, by modulating global network properties. Inspired by those biological observations, we propose simultaneous local and global forms of artificial neuromodulation for continual learning, as shown in Fig.~\ref{fig:neuromodulation_types}. 

Based on the processes described above, we propose a novel architecture-based CL method, \emph{Configurable Random Weight Network}~(CRWN), that utilizes a neuromodulation-inspired neural network for economical and efficient continual learning. Initially, we present two types of artificial neuromodulation and show that their neuromodulatory signals learn how to adjust \emph{long-lasting}/\emph{unchanging} \emph{random} synaptic weights for specific tasks, enabling task-specific learning. The low-dimensional neuromodulatory signals are the only plastic weights of our proposed network and can be represented by a compact vector, and so CRWN training requires significantly fewer learnable parameters and allows for faster running time. 

The rest of the paper is organized as follows. After reviewing related work, we then introduce two forms of artificial neuromodulation for continual learning, \textbf{Global} and \textbf{Local} modulation. We then show how to apply these methods to generate the two main CRWN architectures, \textbf{Fly-circuit-like architecture}~(\textsc{FlyNet}) and \textbf{Neuromodulation architecture}~(\textsc{NeuroModNet}).
%
%
To compare our CRWN approach with the existing CL methods, we conduct experiments using RoationMNIST, PermutationMNIST, Split-CIFAR100, and SplitMini-ImageNet. We show that our approach can reach or outperform the performance of existing approaches with far fewer learnable parameters and a faster running time.
%

\section{Related Work}
\label{sec:related_work}

As discussed above, models for continual learning must be able to learn several tasks sequentially without catastrophic forgetting~(CF)~\cite{mcclelland1995there, thrun1995lifelong}, and proposed approaches for CL can be generally grouped into three categories, including:
\emph{regularization-based methods} that aim to maintain the learned information of past tasks by using sophisticated regularization terms~\cite{chaudhry2020continual, jung2020continual, mirzadeh2020linear}; \emph{replay-buffer-based methods} that utilize and revisit a set of actual or synthesized samples from the previous tasks~\cite{chaudhry2019continual, chaudhry2019tiny, saha2021gradient}; and \emph{architecture-based methods} that minimize the interference between tasks through specially designed architectural components~\cite{cheung2019superposition, wen2020batchensemble, wortsman2020supermasks, kang2022forget}.

Most existing examples of neuromodulation-inspired Artificial Neural Networks~(ANNs) explicitly focus on adjusting critical features of ANNs, such as the slope and bias of the activation or the scale of the weights, or introducing new neuron formalisms that can change the connections between each other~\cite{vecoven2020introducing,tsuda2022neuromodulators,miconi2020backpropamine}. Neuromodulation approaches to the CF problem tend to be related to regularization-based methods. For example, \cite{beaulieu2020learning} use a neuromodulation-inspired ANN architecture that trains two parallel networks concurrently, with one network gating the activations of the second network and thus affecting gradient updates to mitigate CF. \cite{daram2020exploring} use a fly-brain-like network based on context-based DA modulation for one-shot image classification, but neuromodulation is employed for regularization of plasticity in the output layer.

In contrast with other neuromodulation approaches to CF, we employ an architecture-based method, and we compare its performance to other state-of-the-art architecture methods for continual learning~-- \textsc{PSP}, \textsc{BatchE}, and \textsc{SupSup}~-- which do not employ neuromodulation. \textsc{PSP}~\cite{cheung2019superposition} superimposes many models into one by using different (and nearly orthogonal) contexts for each task so the task parameters can be effectively retrieved using the correct task context. \textsc{BatchE}~\cite{wen2020batchensemble} learns a shared weight matrix on the first task and then learns only a rank-one elementwise scaling matrix for each subsequent task. Finally, \textsc{SupSup}~\cite{wortsman2020supermasks} learns a binary mask for each task in a single model so that the model produces a subnetwork of the original network for each task.

\section{Configurable Random Weight Networks~(CRWN)}
\label{sec:crwn}

\subsection{Problem Statement}
We consider an online supervised learning setup where $\mathcal{T}$ tasks arrive to a learner in sequential order. Let $\mathcal{D}_{t} = \{ \textbf{x}_{i, t}, y_{i,t} \}_{t}^{n_{t}}$ be the dataset of task $t$, composed of $n_{t}$ pairs of raw instances and corresponding labels. Our continual learner consists of a neural network $f(\cdot; \boldsymbol\theta_{t})$ for each task~$t$ that is parameterized by the model weights $\boldsymbol\theta_{t}$, and the continual-learning scenario aims to learn the sequence of $\mathcal{T}$ tasks by finding, for each task~$t$, the optimal weights:
\begin{equation}
    \boldsymbol\theta^{*}_{t} = \min_{\boldsymbol\theta_{t}}\frac{1}{n_{t}}\sum_{i=1}^{n_{t}}\mathcal{L}(f(\textbf{x}_{i, t}; \boldsymbol\theta_{t}), y_{i, t})
\label{eq:problem_statement}
\end{equation}
where $\mathcal{L}(\cdot,\cdot)$ is classification loss, such as cross-entropy loss. Dataset $\mathcal{D}_{t}$ for task~$t$ is only accessible when learning task~$t$, and we further assume that task identity is given in both the training and testing stage. As described below, we define two different \emph{Configurable Random Weight Network~(CRWN)} architectures for this problem.

\subsection{\textsc{FlyNet}: Fly-Circuit-like Architecture}
\label{sec:flynet}
In this section, we define \textsc{FlyNet}, a fly-circuit-like single-hidden-layer~(SHL) network for continual learning. \textsc{FlyNet} with $d_{\text{hidden}}$ hidden nodes is a parametric function $f_{d_{\text{hidden}}}: \mathbb{R}^{d_{\text{in}}} \to \mathbb{R}^{d_{\text{out}}}$ taking an input $\textbf{x} \in \mathbb{R}^{d_{\text{in}}}$. The \textsc{FlyNet} architecture includes a randomized \emph{encoder} for projecting the input into a high-dimensional space in a hidden layer and a \emph{decoder} aided by our proposed neuromodulation for continual learning~(described below). The overall \textsc{FlyNet} architecture is depicted in Fig.~\ref{fig:flynet} in the Appendix. 

\paragraph{Encoder} We use the approach of \cite{hong2021insect,hong2021representing} to define the randomized encoder in the hidden nodes. We select $N_{\text{in}} \in [1,d_{\text{in}})$ indices of the input at random without replacement to be gated into each hidden node so that the output $\Bar{x}_j$ of each hidden node $j \in \{1,...,d_{\text{hidden}}\}$ is a sparsely weighted combination of all $d_{\text{in}}$ inputs; that is, only $N_{\text{in}} < d_{\text{in}}$ inputs have $w_{j,i}=1$, and all other inputs have $w_{j,i}=0$. After the $j$-th hidden node produces weighted sum $\Bar{x}_{j}$, the post-processing step produces the intermediate output:
\begin{equation*}
    \hat{x}_{j} \triangleq \Bar{x}_{j} - \gamma\cdot\mu,
    \quad
    \text{ with }
    \quad
    \mu 
    \triangleq
    (1/d_{\text{hidden}})
    \textstyle\sum_{i=1}^{d_{\text{hidden}}} \Bar{x}_{i}
\end{equation*}
where $d_{\text{hidden}}$ is the size of the hidden layer and $\gamma$ is a scalar for controlling the sparsity of the resulting hidden representation\footnote{In neuroscience, the sparsity of representation in the Antennal Lobe and Kenyon Cells is generally considered to be 5$\%$, but we set $\gamma \equiv 1$ for convenience of implementation.}. Then, the sparse output of the $j$-th hidden node can be defined as $h_{j} \triangleq g(\hat{x}_{j})$ where $g$ is the ReLU function that allows the model to produce sparse hidden output. Thus, the hidden output vector is defined as $\textbf{h} \triangleq [h_{1},\dots,h_{d_{\text{hidden}}}]^{\top}$.

\paragraph{Decoder} 
We propose a novel neuromodulation-inspired decoder for CL as depicted in Fig.~\ref{fig:neuromodulation_types}. Given a total $\mathcal{T}$ number of tasks that a model must be able to learn sequentially, let: $\textbf{R} \in \mathbb{R}^{d_{\text{out}} \times d_{\text{hidden}}}$ be a non-learnable, randomized weight matrix initialized by a zero-mean Gaussian distribution; $\textbf{v}_{k} \in \mathbb{R}^{d_{\text{hidden}}}$ be a learnable \emph{context vector}~(\textbf{Local Neuromodulation} in Fig.~\ref{fig:neuromodulation_types}) for the $k$-th task; and $\alpha_{k} \in \mathbb{R}$ be a \emph{random strength factor}, a single constant that controls the strength of randomness in $\textbf{R}$ for the task~$k$~(\textbf{Global Neuromodulation} in Fig.~\ref{fig:neuromodulation_types}). In the following, we denote each of the individual $d_{\text{out}}$ vectors of $\textbf{R}$ so that $\textbf{R}^{\top} \triangleq [ \textbf{r}_{1}^{\top}, \textbf{r}_{2}^{\top}, \dots, \textbf{r}_{d_{\text{out}}}^{\top} ]$.

For each task~$k$, we calculate the vector $\textbf{w}_{(k)i}^{\top} \triangleq \alpha_{k} \cdot (\textbf{r}_{i}^{\top} \odot \textbf{v}_{k})$ for each output~$i \in \{1, \dots, d_{\text{out}}\}$ with $\odot$ indicating element-wise multiplication. Using $\textbf{R} \opneuromodulation \textbf{v}_k$ to denote all element-wise operations occurring between the row vectors of $\textbf{R}$ and $\textbf{v}_{k}$ for task~$k$, we collect these $d_{\text{out}}$ vectors into the neuromodulated weight matrix $\textbf{W}_{(k)} \in \mathbb{R}^{d_{\text{out}} \times d_{\text{hidden}} }$ with $\textbf{W}_{(k)}^{\top} \triangleq [ \textbf{w}_{(k)1}^{\top}, \textbf{w}_{(k)2}^{\top}, \dots, \textbf{w}_{(k)d_{\text{out}}}^{\top} ]$. The mapping from the hidden layer to the output is then:
\begin{equation}
    \textbf{y}_{k} = \textbf{W}_{(k)} \textbf{h} = \alpha_{k} \cdot (\textbf{R} \opneuromodulation \textbf{v}_{k}) \textbf{h}
    \label{eq:neuromodulation}
\end{equation}
where $\textbf{h} \in \mathbb{R}^{d_{\text{hidden}}}$ is the hidden representation by the encoder with input $\textbf{x}_{k}$. That is, learnable $\textbf{v}_k$ and $\alpha_{k}$ modulate the long-lasting randomized connections in $\textbf{R}$ to adapt to the specific task $k$. In our setting, we typically fix the randomly initialized weight matrix $\textbf{R}$ and learn the context vector $\textbf{v}_{k}$ and the single constant $\alpha_{k}$ for each task~$k$ so that each context vector and single scalar impose a unique inductive bias for each specific task.

\paragraph{Parameter-superposition interpretation}
Following~\cite{cheung2019superposition}, we can interpret the \textsc{FlyNet} decoder as the \emph{superposition} of multiple task-specific parameters. In a hypothetical, overparameterized mapping $\textbf{y} \triangleq \textbf{W} \textbf{x}$ between an input $\textbf{x}$ and an output $\textbf{y}$, over-parameterization in $\textbf{W}$ implies only a small sub-space spanned by its rows are relevant for an individual task, and the remaining ``space'' in the matrix could be intelligently used for other tasks without interference. Thus, instead of defining $\textbf{W}_{(1)}, \textbf{W}_{(2)}, \dots, \textbf{W}_{(\mathcal{T})}$ as the set of parameters required for each of the $\mathcal{T}$ tasks separately, \cite{cheung2019superposition} defined a superpositioned parameter $\textbf{W}$ with a task-specific linear transformation $\textbf{C}_{k}^{-1}$ as context for the $k$-th task such that $\textbf{W} \triangleq \sum_{i=1}^{\mathcal{T}} \textbf{W}_{(i)} \textbf{C}_{i}^{-1}$. If the context matrices are carefully chosen, the parameter $\textbf{W}_{(k)}$ for the $k$-th task can be retrieved from $\textbf{W}$ using the context matrix $\textbf{C}_{k}$; in particular, $\hat{\textbf{W}}_{(k)} = \textbf{W}\textbf{C}_{k} = \sum_{i=1}^{\mathcal{T}} \textbf{W}_{(i)}(\textbf{C}_{i}^{-1}\textbf{C}_{k})$ will be a noisy estimate of $\textbf{W}_{(k)}$. Thus, it is possible to interpret $\textbf{W}$ as a memory addressed by each context index. 

For comparison, we can cast our neuromodulation approach into a similar structure:
\begin{equation}
    \textbf{y}_{k} = \mathord{\overbrace{(\alpha_{k} \cdot \textbf{R})}^{\mathbf{W} \text{ for } \alpha_k\equiv 1}} (\mathord{\overbrace{\textbf{v}_{k}}^{\mathbf{C}_k}} \odot \textbf{h})
    \label{eq:superposition}
\end{equation}
where $\textbf{v}_{k} \in \mathbb{R}^{d_{\text{hidden}}}$ acts as context vector, $\alpha_{k} \in \mathbb{R}$ is a scalar that regulates the scale of randomness of $\textbf{R}$, and $\textbf{h} \in \mathbb{R}^{d_{\text{hidden}}}$ is the hidden representation produced by the encoder. Writing out the terms in the sums represented by Eqs~(\ref{eq:neuromodulation}) and~(\ref{eq:superposition}) confirms that the two interpretations are mathematically equivalent. In particular, for both interpretations, $\textbf{y}_{k} = [y_{1}, y_{2}, \dots, y_{d_{\text{out}}}]$ is such that:
\begin{equation*}
    y_{i} = \alpha_{k} \cdot (r_{i1}v_{k1}h_{1} + r_{i2}v_{k2}h_{2} + \dots + r_{id_{\text{hidden}}}v_{kd_{\text{hidden}}}h_{d_{\text{hidden}}})
\end{equation*}
where the $i$-th row in $\textbf{r}_{i}^{\top} = [r_{i1}, \dots, r_{id_{\text{hidden}}}]$, the context vector for $k$-th task $\textbf{v}_{k} = [v_{k1}, \dots, v_{kd_{\text{hidden}}}]^{\top}$ and the hidden representation $\textbf{h} = [h_{1}, \dots, h_{d_{\text{hidden}}}]^{\top}$.
%
If we ignore the scalar $\alpha_{k}$ in Eq.~\ref{eq:superposition}, the key difference between our method and \cite{cheung2019superposition} is that the weight matrix $\textbf{R}$ is non-trainable; that is, only context vector $\textbf{v}_{k}$ is trained in ours whereas both must be learned in the approach of \cite{cheung2019superposition}. Following the work on hetero-associative memory by \cite{kanerva2009hyperdimensional}, our method (as well as \cite{cheung2019superposition}) accesses task-specific models using task-specific \emph{context} information that dynamically \emph{routes} an input towards a specific model retrieved from the superposition in \textbf{R} (or \textbf{W}). However, rather than using contexts to learn new weights in \textbf{W}, our approach is to learn only contexts~($\textbf{v}_{k}$) as a \emph{key} that are used to recombine existing \emph{fixed} parameters in non-learnable \textbf{R}. Also, because we utilize the random matrix, the learnable random strength factor $\alpha_{k}$ is introduced to add an additional degree of freedom to further improve learning. Consequently, our approach can drastically reduce spatial and learning complexity and retrieve learned weights entirely without contamination by learning other tasks. 

\subsection{\textsc{NeuroModNet}: Neuromodulation Architecture}
\label{sec:neuromodnet}
The single-hidden-layer architecture of \textsc{FlyNet} was developed to maximize biological congruence. However, the neuromodulatory approach in the \textsc{FlyNet} decoder can be extended as a module that can be used in a wider range of continual learning applications with deep neural networks as well. In particular, the proposed neuromodulation can be applied to the network's input layer so that it can be similarly modulated to improve performance with a much simpler architecture. In this section, we describe \textsc{NeuroModNet} where both input and hidden layers have their own neuromodulatory signals to learn. The overall architecture of \textsc{NeuroModNet} is depicted in Fig.~\ref{fig:neuromodnet} in the Appendix.
\begin{table*}[t!]
\caption{\emph{Experimental results of all models on variants of MNIST}. \textbf{R-MNIST Acc.} and \textbf{P-MNIST Acc.} mean average test accuracy after learning all tasks for RotationMNIST and PermutationMNIST, respectively. \textbf{$\#$ Params/Task} indicates the number of learnable parameters per task. \textbf{Ratio of Params} is the ratio of the number of learnable parameters compared to MTL.}
\centering
\resizebox{0.9\textwidth}{!}{%
\begin{tabular}{c|cccc}
\toprule
\textbf{Methods} & \textbf{R-MNIST Acc. ($\%$)} & \textbf{P-MNIST Acc. ($\%$)} & \textbf{$\#$ Params/Task} & \textbf{Ratio of Params} \\
\hline
\hline
MTL, FC256 & $97.18_{\pm 0.06}$ & $97.20_{\pm 0.03}$ & $269$K & $1:1$ \\
\hline
\textsc{PSP}~\cite{cheung2019superposition}, FC256 & $96.16_{\pm 0.06}$ & $93.34_{\pm 0.24}$ & $1$M & $3.77:1$ \\
\textsc{BatchE}~\cite{wen2020batchensemble}, FC256 & $89.21_{\pm 0.17}$ & $89.12_{\pm 0.03}$ & $9$K & $1:28.9$ \\
\textsc{SupSup}~\cite{wortsman2020supermasks}, FC256 & $94.22_{\pm 0.03}$ & $94.23_{\pm 0.01}$ & $269$K & $1:1$ \\
\hline
\textsc{FlyNet}, 10d & $94.18_{\pm 0.02}$ & $94.18_{\pm 0.01}$ & $8$K & $1:34.3$ \\
\textsc{FlyNet}, 20d & $94.74_{\pm 0.07}$ & $94.70_{\pm 0.06}$ & $16$K & $1:17.1$ \\ 
\textsc{FlyNet}, 30d & $94.90_{\pm 0.13}$ & $94.93_{\pm 0.02}$ & $24$K & $1:11.4$ \\ 
\textsc{FlyNet}, 40d & $94.84_{\pm 0.10}$ & $94.93_{\pm 0.06}$ & $31$K & $1:8.6$ \\
\hline
\textsc{NeuroModNet}, FC256 & $90.59_{\pm 0.06}$ & $90.61_{\pm 0.04}$ & $1$K & $1:207$ \\
\textsc{NeuroModNet}, FC512 & $92.14_{\pm 0.05}$ & $92.19_{\pm 0.04}$ & $2$K & $1:148$ \\
\textsc{NeuroModNet}, FC1024 & $93.51_{\pm 0.06}$ & $93.56_{\pm 0.04}$ & $3$K~ & $1:94.8$ \\
\textsc{NeuroModNet}, FC2048 & $94.62_{\pm 0.08}$ & $94.65_{\pm 0.01}$ & $5$K & $1:55$ \\
\textsc{NeuroModNet}, FC4096 & $95.49_{\pm 0.06}$ & $95.53_{\pm 0.05}$ & $9$K & $1:29.9$ \\
\bottomrule
\end{tabular}%
}
\label{tab:results_variants_mnist}
\end{table*}
\paragraph{Application to Multi-layer Neural Networks} 
We can extend the above formulations to the entire neural network models by applying Eqs~(\ref{eq:neuromodulation}) and~(\ref{eq:superposition}) to the linear transformations of all layers $l$ of a neural network:
\begin{align*}
    \textbf{x}^{(l+1)}
    &= g(\alpha_{k} \cdot (\textbf{R} \opneuromodulation \textbf{v}_{k}^{(l)}) \textbf{x}^{(l)})\\
    &= g((\alpha_{k} \cdot \textbf{R}) (\textbf{v}_{k}^{(l)} \odot \textbf{x}^{(l)}))
\end{align*}
where $g(\cdot)$ is a non-linear activation, such as ReLU. 

\paragraph{Application to Convolutional Networks} For vision tasks, convolution is conventionally the dominant operation performed in most layers of neural-network models. In this setting, we apply a context tensor $\textbf{v}_{k} \in \mathbb{R}^{M \times H_{w} \times W_{w}}$ to the non-learnable convolution kernel $\textbf{R} \in \mathbb{R}^{N \times M \times H_{w} \times W_{w}}$ with the input image $x \in \mathbb{R}^{M \times H_{w} \times W_{w}}$ as follows:
\begin{equation*}
    y_{n} = \alpha_{k} \cdot (\textbf{R}_{n} \odot \textbf{v}_{k}) * x
\end{equation*}
where $*$ is convolution operation, $M$ is the dimension of the input channel, and $N$ is the dimension of the output channel.

\section{Experiments}
\label{sec:exp}

\begin{figure*}[t]
\centering
\includegraphics[width=0.98\textwidth]{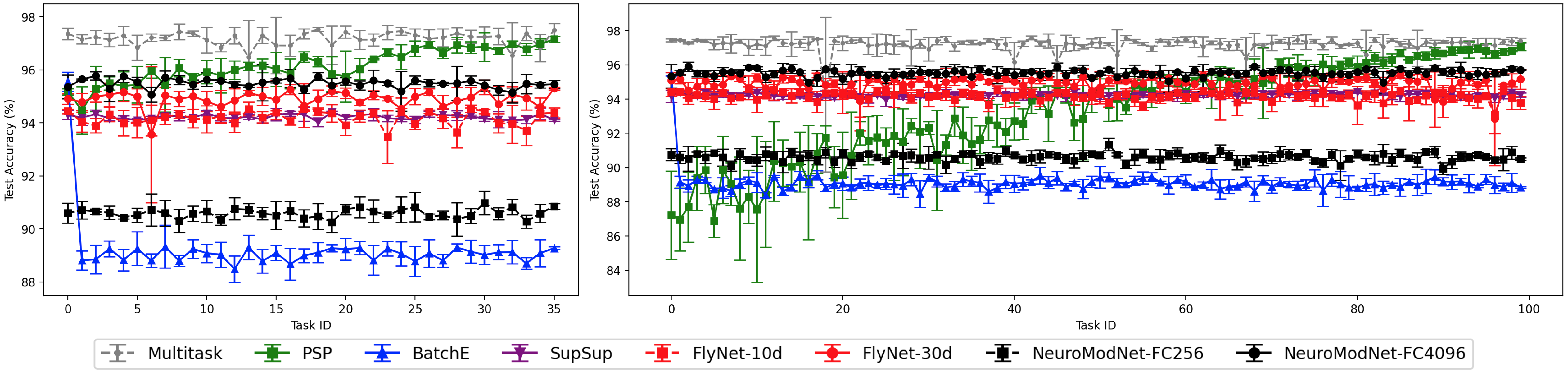}
\caption{Test accuracy of all tasks for RotationMNIST~(\textbf{left}) and PermutationMNIST~(\textbf{right}) after the completion of learning all tasks. Despite far fewer adaptable parameters, our method, \textsc{FlyNet} and \textsc{NeuroModNet}, achieves similar or better performances than other existing methods.}
\label{fig:exp_mnist_01}
\end{figure*}

In this section, we validate the CRWN method through experiments demonstrating continual learning performance on various benchmarks. 
We follow the experimental setups in recent works~\cite{cheung2019superposition,chaudhry2019tiny, wortsman2020supermasks}. All experimental results in the figures and tables are averaged over three trials with different random seeds.

\paragraph{Datasets and Architectures} We use four different popular CL benchmarks as follows:
\begin{itemize}
    \item \textbf{RotationMNIST~(R-MNIST)}~\cite{wortsman2020supermasks}, a variant of MNIST~\cite{lecun1998mnist} where each of 36 tasks corresponds to images rotated by a multiple of 10 degrees
    \item \textbf{PermutedMNIST~(P-MNIST)}~\cite{kirkpatrick2017overcoming}, another variant of MNIST where new tasks are created with a fixed random permutation of the pixels of MNIST, and the total number of tasks in the experiment is 100
    \item \textbf{Split-CIFAR100}~\cite{wen2020batchensemble} randomly partitions CIFAR100~\cite{krizhevsky2009learning} into 20 different 5-way classification problems
    \item \textbf{Split-miniImageNet}~\cite{chaudhry2019tiny} splits the Mini-ImageNet dataset~\cite{vinyals2016matching}, a subset of ImageNet~\cite{deng2009imagenet} with 100 classes, into 20 disjoint tasks where each task contains 5 classes and every class consists of 500 $3 \times 84 \times 84$ images for training and 100 images for testing
\end{itemize}

For experiments on R-MNIST and P-MNIST, we use a two-hidden-layer fully connected~(FC) network of 256 units per hidden layer~(FC256) for other baseline methods. Because our \textsc{FlyNet} is a single-hidden-layer model, it uses different sizes of hidden layers by multiplying the input size by the scaling factors $\{ 10d, 20d, 30d, 40d \}$~(i.e., $10d$ means the input dimension $\times 10$). For \textsc{NeuroModNet}, we use a two-hidden-layer FC network with the set of unit sizes per hidden layer, $\{ 256, 512, 1024, 2048, 4096 \}$, to show the trade-off between the number of training parameters and performance. For experiments on Split-CIFAR100 and Split-miniImageNet, we use a reduced ResNet-18 and a standard ResNet-18~\cite{he2016deep}, respectively, for other baseline models. \textsc{NeuroModNet} is the only CRWN that can be applied for experiments on Split-CIFAR100 and Split-miniImageNet. For Split-CiFAR100, we apply \textsc{NeuroModNet} to the same reduced ResNet-18 as other baseline methods but with the multiplication of output channels with the scaling factors, $\{ 1d, 2d, 4d, 6d \}$, showing the relationship between the number of learnable parameters and performance. For Split-miniImageNet, \textsc{NeuroModNet} is applied to a standard ResNet-18.

\paragraph{Baselines}
We compare our CRWN with the recent CL baselines~-- \textbf{PSP}, \textbf{BatchE}, and \textbf{SupSup}. In summary:
\begin{itemize}
    \item \textbf{PSP}~\cite{cheung2019superposition} superimposes many models into one using different contexts for each task. The task parameters can then be effectively retrieved using the correct task context.
    \item \textbf{BatchE}~\cite{wen2020batchensemble} learns a shared weight matrix on the first task and then learns only a rank-one element-wise scaling matrix for each subsequent task. Because learning the rank-one matrix allows for high computational efficiency, and we regard this method as a corresponding efficiency baseline.
    \item \textbf{SupSup}~\cite{wortsman2020supermasks} finds supermasks~(subnetworks) within a randomly initialized network for each task in continual learning. Because this approach finds subnetworks using a search space equal to the number of parameters in the original dense network, we consider it as a baseline focused on performance.
    \item \textbf{FlyNet}~(ours) is a single-hidden-layer neural network that learns a context vector and a single scalar for continual learning.
    \item \textbf{NeuroModNet}~(ours) generalizes the proposed neuromodulation for continual learning so that it performs well in more complex experimental settings.
\end{itemize}
We also compare to non-CL methods \textbf{Multitask Learning~(MTL)} and \textbf{Single-task Learning~(STL)} (MTL trains on multiple tasks simultaneously, and STL trains on a single task independently).
\begin{figure}[t!]
\centering
\includegraphics[width=0.45\textwidth]{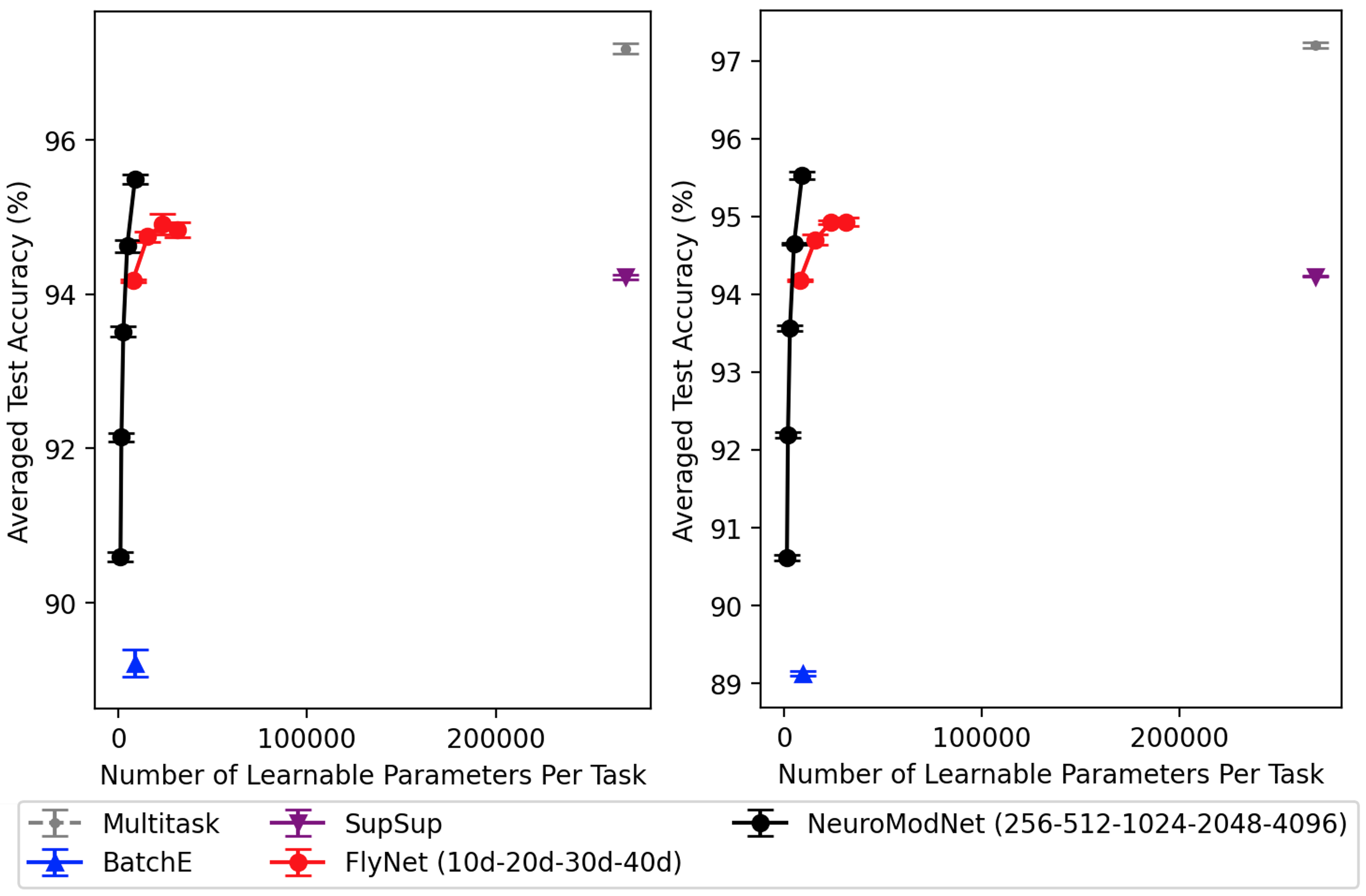}
\caption{Averaged test accuracy over all tasks for RotationMNIST~(\textbf{left}) and PermutationMNIST~(\textbf{right}). \textsc{PSP} is omitted because the model requires much larger adaptable parameters than the multitask model~(See Table~\ref{tab:results_variants_mnist}). }
\label{fig:exp_mnist_02}
\end{figure}
\begin{table*}[t!]
\caption{\emph{Experimental results of all models on Split-CIFAR100 and Split-miniImageNet}. \textbf{Acc.} means average test accuracy after learning all tasks. \textbf{$\#$ Params/Task} indicates the number of learnable parameters per task. \textbf{Ratio of Params} is the ratio of the number of learnable parameters between each model and STL. \textbf{Time} is the average running time per task, including training and inference time~(250 epochs), and each model's configuration time~(\textbf{Time} for \textsc{PSP} is ignored due to its training failure).}
\centering
\resizebox{\textwidth}{!}{%
\begin{tabular}{c|cccc|cccc}
\toprule
\textbf{Methods} & \multicolumn{4}{c}{\textbf{Split-CIFAR100}} & \multicolumn{4}{c}{\textbf{Split-miniImageNet}} \\
\hline
& \textbf{Acc. ($\%$)} & \textbf{$\#$ Params/Task} & \textbf{Ratio of Params} & \textbf{Time~(sec)} & \textbf{Acc. ($\%$)} & \textbf{$\#$ Params/Task} & \textbf{Ratio of Params} & \textbf{Time~(sec)} \\
\hline
\hline
STL & $89.68_{\pm 0.21}$ & 275K & $1:1$ & - & $76.81_{\pm 0.23}$ & 11M & $1:1$ & - \\
\hline
\textsc{PSP}~\cite{cheung2019superposition} & $25.25_{\pm 1.07}$ & 273K & $1:1.01$ & - & $23.76_{\pm 0.25}$ & 11M & $1:1$ & - \\
\textsc{BatchE}~\cite{wen2020batchensemble} & $72.45_{\pm 0.65}$ & 20K & $1:13.7$ & $414_{\pm 4.6}$ & $67.94_{\pm 0.17}$ & 568K & $1:20$ & $658_{\pm 20.3}$ \\
\textsc{SupSup}~\cite{wortsman2020supermasks} & $86.31_{\pm 0.13}$ & 273K & $1:1.01$ & $249_{\pm 2.1}$ & $73.94_{\pm 0.05}$ & 11M & $1:1$ & $820_{\pm 3.4}$ \\
\hline
\textsc{NeuroModNet}, 1d & $81.66_{\pm 0.39}$ & 5K & $1:55$ & $\mathbf{243_{\pm 0.8}}$ & $72.69_{\pm 0.89}$ & 55K & $1:203$ & $\mathbf{564_{\pm 1.6}}$ \\
\textsc{NeuroModNet}, 2d & $84.81_{\pm 0.91}$ & 10K & $1:28$ & - & - & - & - & - \\
\textsc{NeuroModNet}, 4d & $85.98_{\pm 0.19}$ & 20K & $1:14$ & - & - & - & - & - \\
\textsc{NeuroModNet}, 6d & $86.56_{\pm 0.28}$ & 30K & $1:9$ & - & - & - & - & - \\
\bottomrule
\end{tabular}%
}
\label{tab:results_splitcifar100_splitminiimagenet}
\end{table*}
\paragraph{Experimental Settings} We directly implement our method from the official code of~\cite{wortsman2020supermasks} and utilize the implementation of \textsc{PSP}, \textsc{BatchE}, and \textsc{SupSup} from it. For Split-CIFAR100 and Split-miniImageNet, we deploy the convolution version of \textsc{PSP} from the official code of~\cite{cheung2019superposition} and compare our approach with it. We provide more details of the datasets, architectures, hardware specifications, and experimental settings, including the hyperparameter configurations for all methods, in the Appendix.
\begin{figure*}[t!]\centering%
\begin{subfigure}[t]{0.48\textwidth}\centering%
\includegraphics[width=\textwidth]{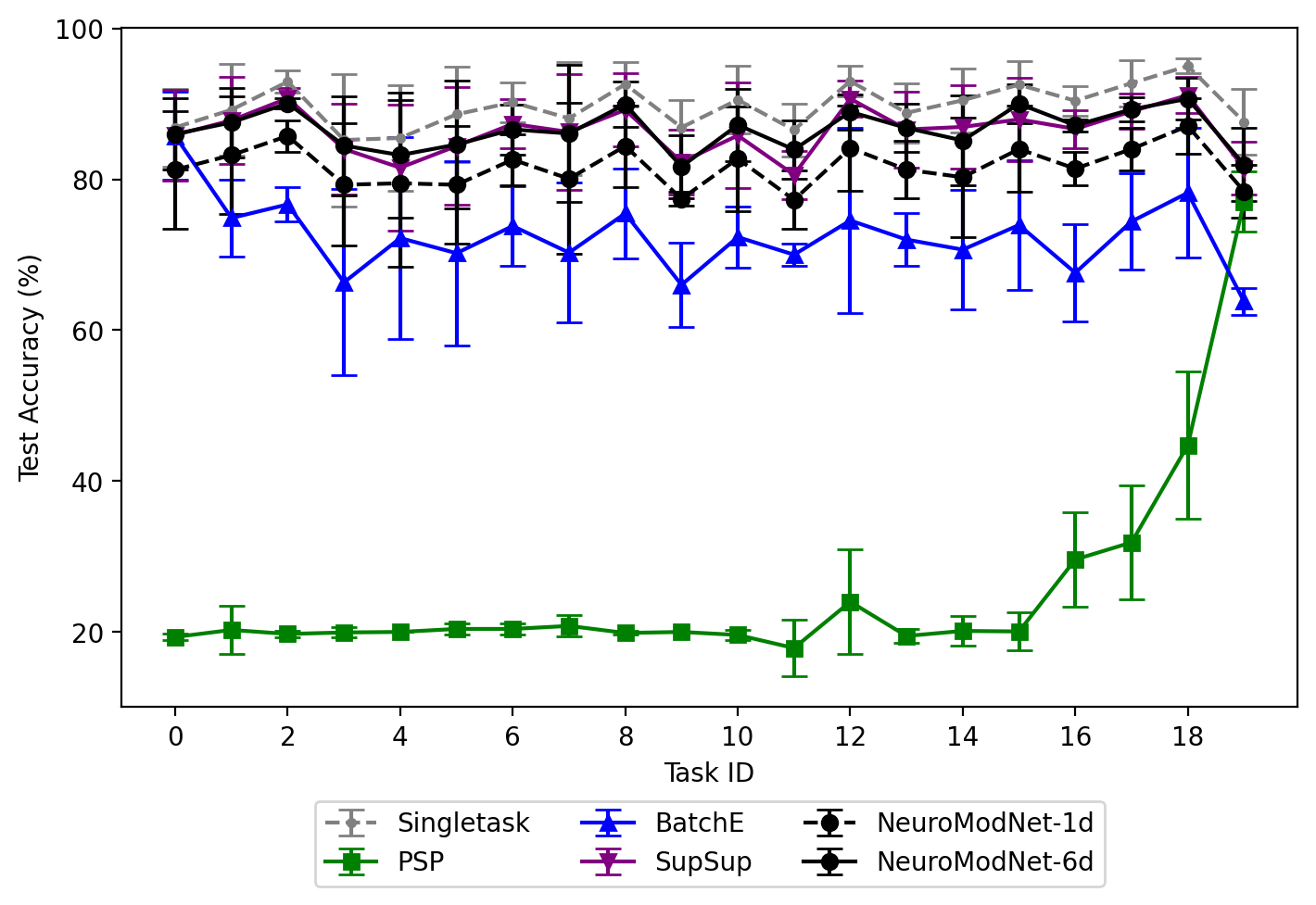}
\caption{Testing each model on Split-CIFAR100.}
\label{fig:splitcifar100_testacc_taskid}
\end{subfigure}
\hfill
\begin{subfigure}[t]{0.48\textwidth}\centering%
\includegraphics[width=\textwidth]{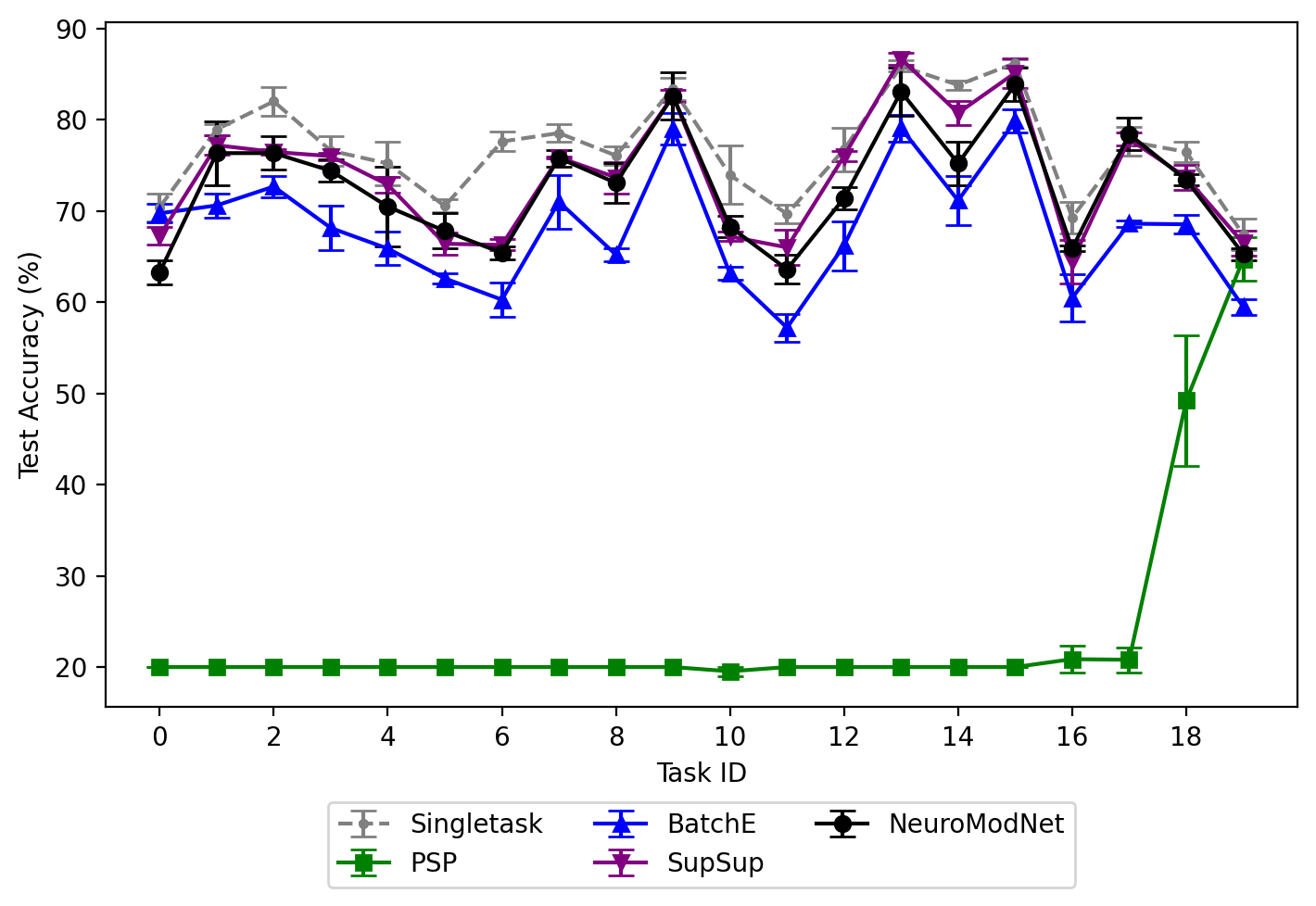}
\caption{Testing each model on Split-miniImageNet.}
\label{fig:splitcifar100_testacc_numparam}
\end{subfigure}
\hfill
\caption{Test accuracy of all tasks for Split-CIFAR100 (\textbf{left})
and Split-miniImageNet (\textbf{right}) after learning all tasks.}
\label{fig:splitcifar100_splitminiimagenet_01}
\end{figure*}
\begin{figure}[t!]
\centering
\includegraphics[width=0.47\textwidth]{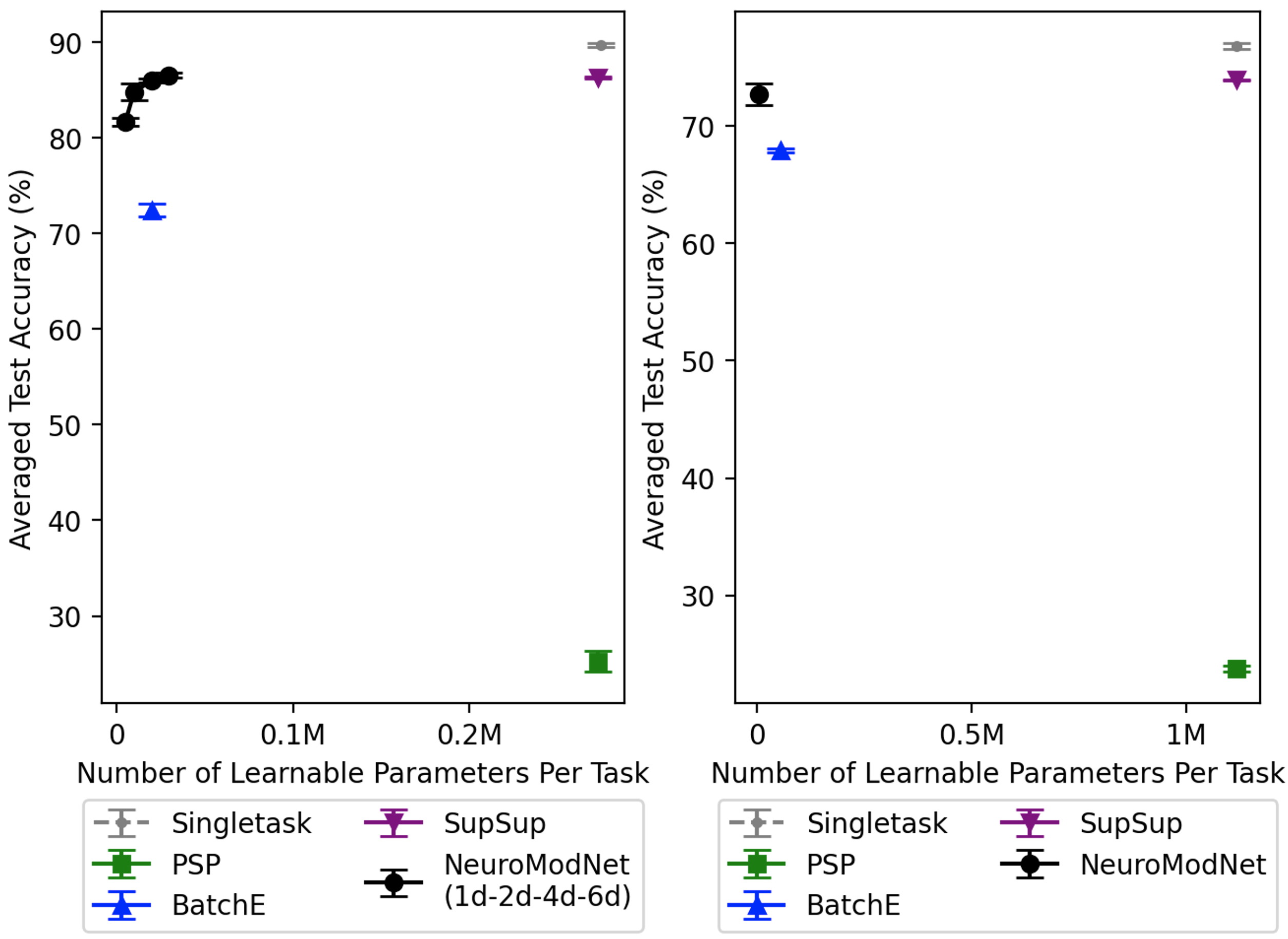}
\caption{Average Test accuracy of all tasks for Split-CIFAR100 (\textbf{left})
and Split-miniImageNet (\textbf{right}).}
\label{fig:splitcifar100_splitminiimagenet_02}
\end{figure}

\paragraph{Results on MNIST variants}
Table~\ref{tab:results_variants_mnist} shows the overall experimental results of all models on R-MNIST and P-MNIST. In the case of \textsc{PSP}, there is a performance gap between R-MNIST~($96.16$) and P-MNIST~($93.34$) because the weight matrix of PSP is constantly changing to learn subsequent tasks, which is a limitation of the method that becomes more apparent as the number of tasks to be learned increases. Furthermore, due to the architectural nature of the \textsc{PSP}, it requires much more learnable parameters~(1M) than the non-CL MTL method~(269K). For our CRWN, our \textsc{NeuroModNet}-FC256 outperforms \textsc{BatchE} by achieving $90.59$ and $90.61$ for R-MNIST and P-MNIST, respectively, with far fewer learnable parameters, which shows that our approach is more computationally efficient than \textsc{BatchE}. In terms of performance, we show that the performance of \textsc{NeuroModNet} can easily reach or surpass that of \textsc{SupSup} by utilizing relatively small spatial complexity. Specifically, noting the ratio of training parameters between the non-CL method and the model, our \textsc{NeuroModNet}-FC4096 achieves the best performance by utilizing a significantly smaller number of training parameters at some spatial complexity. On the other hand, despite the limitation that \textsc{FlyNet} is a single-hidden-layer architecture, \textsc{FlyNet}-10d performs much better than \textsc{BatchE} in terms of performance and efficiency. Similar to \textsc{NeuroModNet}, \textsc{FlyNet} can outperform \textsc{SupSup} with a far fewer number of adaptable parameters at the expense of some spatial complexity.

Figure~\ref{fig:exp_mnist_01} shows the test accuracy of all tasks for R-MNIST and P-MNIST after learning all tasks. Compared to other models, \textsc{PSP} shows the limitation that the achieving performances are inconsistent. This is because the structural nature of \textsc{PSP} keeps updating its main weight for learning subsequent tasks. Figure~\ref{fig:exp_mnist_02} depicts the average test accuracy across all tasks for R-MNIST and P-MNIST with the number of learnable parameters per task for each model. Our CRWN, both \textsc{FlyNet} and \textsc{NeuroModNet}, outperforms \textsc{BatchE} and can reach or surpass the performance of \textsc{SupSup} by leveraging additional special complexity.

\paragraph{Results on Split-CIFAR100 and Split-miniImageNet}
Table~\ref{tab:results_splitcifar100_splitminiimagenet} shows the overall experimental results of all models on Split-CIFAR100 and Split-miniImageNet. In our CRWN, we only show results for \textsc{NeuroModNet} as our \textsc{FlyNet} cannot be applied to these problems. \textsc{PSP} performs poorly on both problems, revealing the model's inherent limitations. On the other hand, our \textsc{NeuroModNet} outperforms \textsc{BatchE} in terms of computational efficiency with a smaller number of learnable parameters and faster execution time per task. Specifically, in the experiments on Split-miniImageNet, although our \textsc{NeuroModNet} is slightly worse than the performance of \textsc{SupSup}, it shows that our approach indeed possesses faster running time and efficient computation. See Appendix for additional figures.

As seen from the experimental results, our model is a novel, computationally efficient approach for continual learning, but its significant decrease in number of learnable parameters and training time comes with limitations in terms of performance. Furthermore, to achieve state-of-the-art performance, our method requires additional spatial complexity, which may be burdensome in specific environments. Thus, our approach may be ideal for environments that are focused on high computational efficiency while maintaining good performance.

Figure~\ref{fig:splitcifar100_splitminiimagenet_01} demonstrates test accuracy over all tasks for Split-CIFAR100 and Split-miniImageNet after completing learning all tasks. Compared to other baselines, \textsc{PSP} failed to achieve reasonable performances for both experiments as a CL method by suffering catastrophic forgetting. 

Figure~\ref{fig:splitcifar100_splitminiimagenet_02} shows average test accuracy of all tasks for Split-CIFAR100 and Split-miniImageNet. In terms of both computation efficiency and performance, our \textsc{NeuroModNet} can perform reasonably compared to other baselines.

\paragraph{Ablation} 

\begin{table}[t!]
\caption{\emph{Ablation study on R-MNIST to show the effectiveness of learning global and local modulations.} For each configuration of CRWN, average test accuracy of all tasks after learning all tasks is shown in the table.}
\centering
\begin{tabular}{c|c|c}
\toprule
\textbf{Model} & \textbf{Local + Global} & \textbf{Local Only} \\
\hline
\hline
\textsc{FlyNet}-10d & $94.18_{\pm 0.02}$ & $93.87_{\pm 0.03}$ \\
\textsc{FlyNet}-20d & $94.74_{\pm 0.07}$ & $94.51_{\pm 0.03}$ \\ 
\textsc{FlyNet}-30d & $94.90_{\pm 0.13}$ & $94.77_{\pm 0.03}$ \\ 
\textsc{FlyNet}-40d & $94.84_{\pm 0.10}$ & $94. 93_{\pm 0.02}$ \\
\hline
\textsc{NeuroModNet}-FC256 & $90.59_{\pm 0.06}$ & $90.12_{\pm 0.09}$ \\
\textsc{NeuroModNet}-FC512 & $92.14_{\pm 0.05}$ & $91.63_{\pm 0.07}$ \\
\textsc{NeuroModNet}-FC1024 & $93.51_{\pm 0.06}$ & $92.97_{\pm 0.02}$ \\
\textsc{NeuroModNet}-FC2048 & $94.62_{\pm 0.08}$ & $94.04_{\pm 0.02}$ \\
\textsc{NeuroModNet}-FC4096 & $95.49_{\pm 0.06}$ & $94.79_{\pm 0.02}$ \\
\bottomrule
\end{tabular}
\label{tab:modulation}
\end{table}
In Table~\ref{tab:modulation}, we present results of an ablation study on our CRWN using R-MNIST to illustrate the relative contribution of global neuromodulation~(i.e., $\alpha_{k}$ in Eqs.~(\ref{eq:neuromodulation}) and~(\ref{eq:superposition})) and local neuromodulation~(i.e., $\textbf{v}_{k}$ in Eqs.~(\ref{eq:neuromodulation}) and~(\ref{eq:superposition})). Using global neuromodulation only is unlikely to perform well due to the low degrees of training freedom, and so we compare performance of the two modulations together to the performance of local neuromodulation only. Although the performance of both neuromodulations together is better than local neuromodulation alone, learning global modulation affects \textsc{FlyNet} and \textsc{NeuroModNet} differently. For \textsc{NeuroModNet}, learning global modulation consistently contributes to performance regardless of the size of the network. On the other hand, for \textsc{FlyNet}, as the model size increases, global modulation's efficiency gradually decreases, and learning of global modulation may degrade performance, indicating that hidden layers of a specific size or larger in \textsc{FlyNet} show that learning of local modulation dominates over learning of global modulation. We provide an additional ablation study in the Appendix that shows the efficiency of using random, fixed weights in CRWN.

\begin{figure}[t!]
    \centering
    \includegraphics[width=0.4\textwidth]{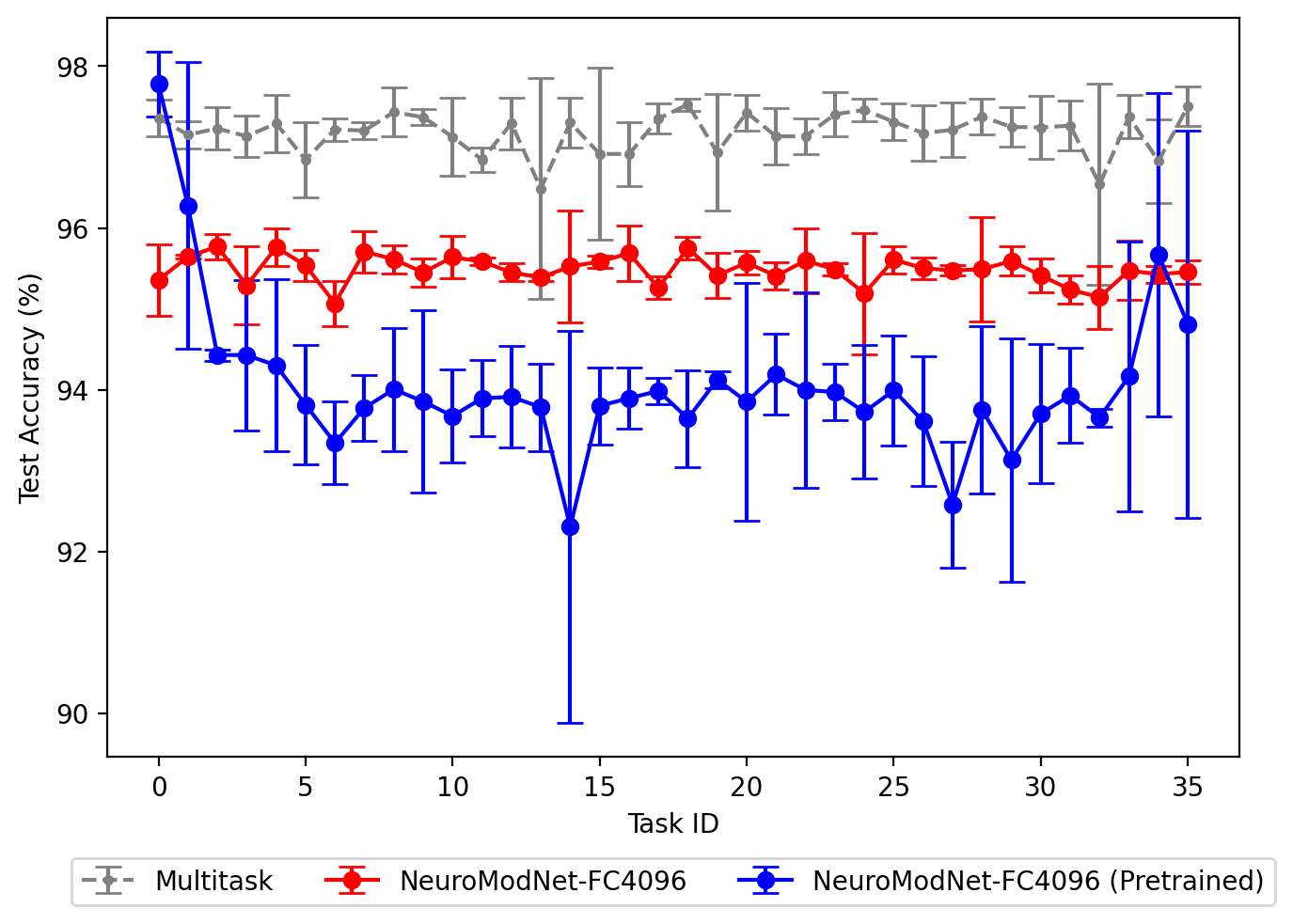}
    \caption{Comparison of \textsc{NeuroModNet} with and without pretrained mode.}
    \label{fig:rmnist_neuromodnet_pretrain}
\end{figure}

We present an ablation study of our CRWN on R-MNIST to demonstrate the efficiency of using random weights acting as a general feature extractor. We show that utilizing fixed random weights as a shared weight can be suitable for continual learning. Inspired by~\textsc{BatchE}~\cite{wen2020batchensemble}, we apply the \emph{pretrained} mode to our \textsc{NeuroModNet}~-- learning a shared weight matrix on the first task and then learning both global and local modulations for each subsequent task based on the shared weight. Figure~\ref{fig:rmnist_neuromodnet_pretrain} depicts the difference in the performance of models with and without the pretrained mode. The model with the pretrained mode performs better on the first two tasks, but the overall performance without the pretrained model is consistently better. 

Furthermore, Table~\ref{tab:rmnist_neuromodnet_pretrain} shows the difference in performances of the model with and without the pretrained model. Intriguingly, the model with the pretrained mode performs worse than the one without the model despite of utilizing extremely larger computational resources. This shows that the pretrained shared weight can be seen as the transferred knowledge without sophisticated guidance to our model and can act as a kind of \emph{prejudice} that severely prevents learning subsequent tasks. Thus, in this case, utilizing \emph{uninformative} random weights can be beneficial. 
\begin{table}[t!]
\caption{Average test accuracy of \textsc{NeuroModNet} with pretrained mode and without pretrained mode. \textbf{Average Acc.} means the average test accuracy of all tasks on R-MNIST. \textbf{$\#$ Params/Task} indicates the number
of learnable parameters per task.}
\centering
\begin{tabular}{c|c|c}
\toprule
\textbf{Mode} & \textbf{Average Acc.~($\%$)} & \textbf{$\#$ Params/Task} \\
\hline
\hline
\textbf{w/} Pretrained & $94.05_{\pm 0.35}$ & 565K \\
\textbf{w.o/} Pretrained & $\boldsymbol{95.49}_{\pm 0.06}$ & \textbf{9K} \\ 
\bottomrule
\end{tabular}
\label{tab:rmnist_neuromodnet_pretrain}
\end{table}

\section{Discussion}
\label{sec:discussion}
We have presented a fundamentally different approach for economically addressing catastrophic forgetting through artificial neuromodulation, learning to regulate a fixed weight matrix using task-specific context vectors and random strength factors for multiple tasks. Our model is compatible with fully connected and convolutional nets, showing its applicability to state-of-the-art networks like ResNet, and our CL method is the first whose learned parameters can be used to provide semantic representations among different tasks.

Our work can be extended in several ways. First, it can be extended with biologically plausible learning methods, potentially informing methods for tuning the randomized fixed-weight matrix. Furthermore, it can be applied to various domains, such as representing superpositions of many rules in neuro-symbolic computing or novel applications of compact context vectors for privacy-preserving and federated learning.

{\small
\bibliographystyle{unsrt}
\bibliography{main}
}

\clearpage
\appendix
\section*{Appendix}
\label{sec:appendix}
\renewcommand\thefigure{A-\arabic{figure}}\setcounter{figure}{0}
\renewcommand\thetable{A-\arabic{table}}
\setcounter{table}{0}
\setcounter{section}{0}
\pagenumbering{Alph}

\section{Architecture Details of Configurable Random Weight Network~(CRWN)}

\subsection{Fly Circuit as Efficient Computational Model} 
In the fruit-fly brain, the simple three-layer circuit from the Antennal Lobe~(AL) to the Kenyon Cells~(KCs) to the Mushroom Body Output Neurons~(MBONs) has been shown to play a critical role in learning to associate odors with electric shocks or sugar rewards and adapting behaviors conveyed by reinforcement signals~\cite{modi2020drosophila}. Due to its capabilities and simple relatively feed-forward structure, the AL--KC--MBON circuit has been utilized as an efficient computational model in computer science and machine learning as the form of implementing a  single-hidden-layer neural network~\cite{dasgupta2017neural,dasgupta2018neural,liang2021can,sinha2021fruit,hong2021insect,hong2021representing}.

\subsection{Biological Neuromodulation}
Neuromodulation has been shown to play an essential role in learning and adaptive behavior in humans and insects and to operate on spatio-temporal scales that reconfigure local neural networks' functions and regulate global cognitive states~\cite{mei2022informing}. Furthermore, neuromodulation has been studied extensively in the insect brain, which can remap neuronal representations conveyed by dopaminergic signals~\cite{modi2020drosophila}. 
In particular, due to the dopaminergic signals, the fly circuit can act as a \emph{switchboard} to include adaptive control of behavior based on internal state and environmental context~\cite{modi2020drosophila}. Switchboarding in the insect brain involves state- and context-dependent Dopaminergic Neuron~(DAN) activity that modulates signal transmission between KCs and MBONs, routing the same sensory input to different output channels. According to this scheme, signal transmission through the fly circuit is modulated moment-by-moment by DAN activity~\cite{modi2020drosophila}. 

Neuromodulation has attracted little attention in computer science or the machine-learning literature. Therefore, we combine the two main ideas of biological neuromodulation above~(See Fig.~\ref{fig:neuromodulation}) and present our initial CL method by implementing a fly-circuit-like single-hidden-layer neural network, \textsc{FlyNet}, that adapts to behavior contexts by dopaminergic signals. In addition, we can generalize the proposed neuromodulation to \textsc{NeuroModNet} to mimic the adaptation of contexts at all layers.

\subsection{Overall Architecture of CRWN}
\begin{figure*}[t!]\centering%
\begin{subfigure}[t]{0.5\textwidth}\centering%
\includegraphics[width=\textwidth]{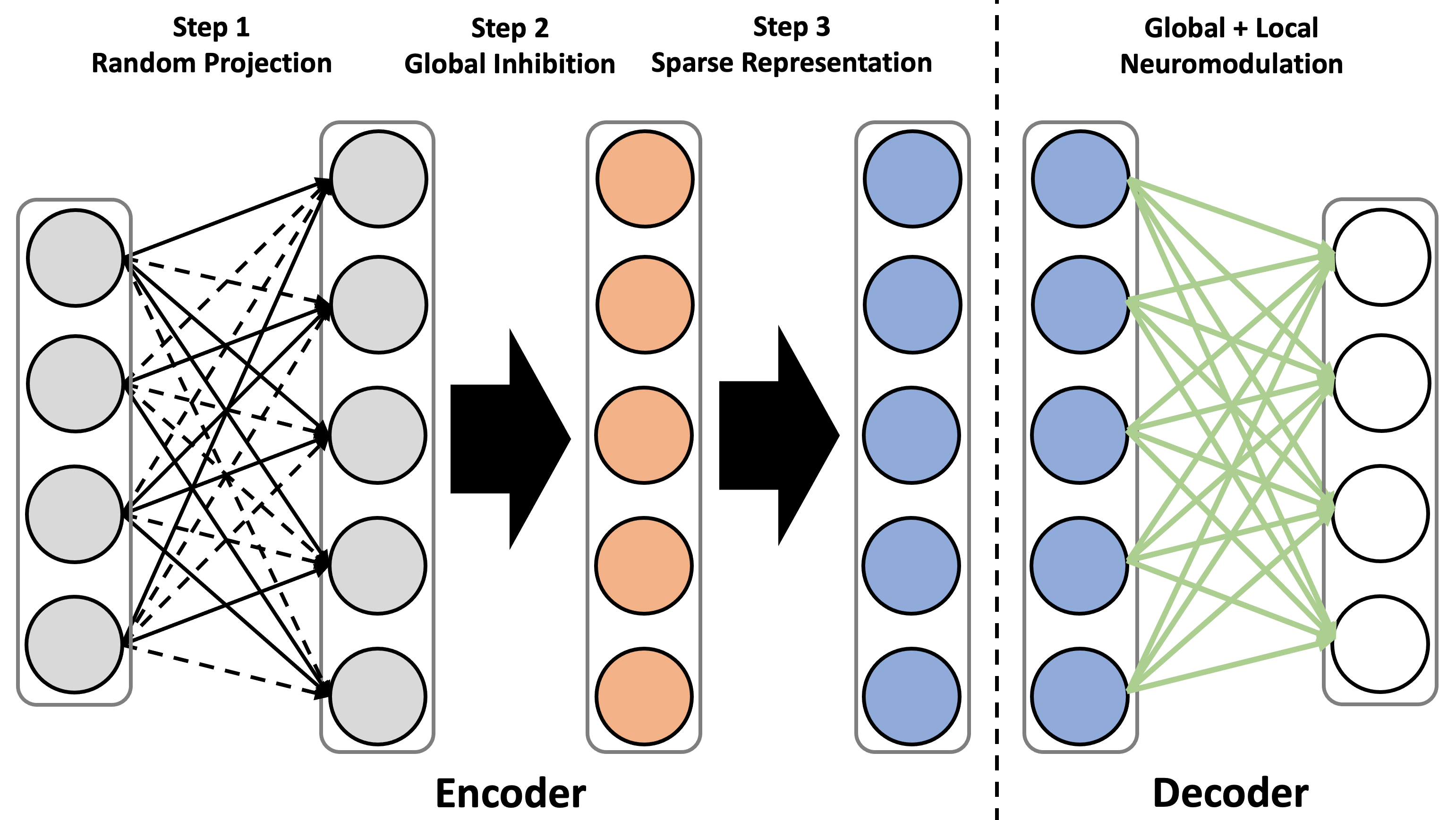}
\caption{Architecture of \textsc{FlyNet}}
\label{fig:flynet}
\end{subfigure}
\hfill
\begin{subfigure}[t]{0.4\textwidth}\centering%
\includegraphics[width=\textwidth]{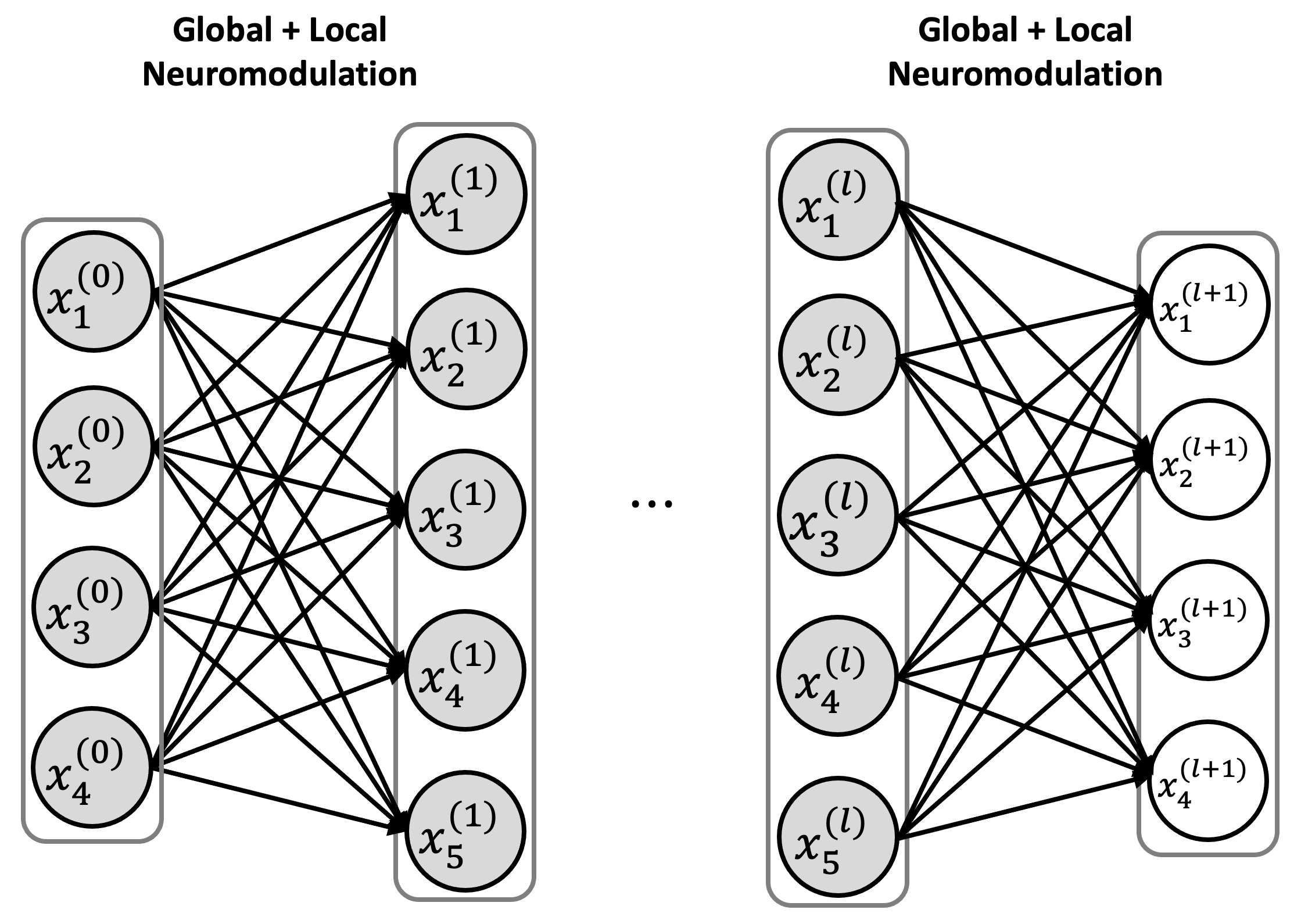}
\caption{Architecture of \textsc{NeuroModNet}}
\label{fig:neuromodnet}
\end{subfigure}
\hfill
\caption{Overall architecture of Configurable Random Weight Network (CRWN) approaches}
\label{fig:model_crwn}
\end{figure*}
Figure~\ref{fig:model_crwn} depicts the overall architecture of our proposed models. 

\section{Experimental Details}
We describe the details of two experiments: i) for continual learning, and ii) for interpretability. All source codes are available at~\url{https://github.com/jyhong0304/CRWN_CL}.

\subsection{Hardware Specification of The Server}
The hardware specification of the server that we used to experiment is as follows:
\begin{itemize}
    \item CPU: Intel\textregistered{} Core\textsuperscript{TM} i7-6950X CPU @ 3.00GHz (up to 3.50 GHz)
    \item RAM: 128 GB (DDR4 2400MHz)
    \item GPU: NVIDIA GeForce Titan Xp GP102 (Pascal architecture, 3840 CUDA Cores @ 1.6 GHz, 384 bit bus width, 12 GB GDDR G5X memory)
\end{itemize}

\subsection{Experiments for Continual Learning}
We validate our method on several benchmark datasets against relevant continual learning baselines. We follow the experimental setups in recent works~\cite{cheung2019superposition,chaudhry2019tiny, wortsman2020supermasks}. All experimental results in the figures and tables are averaged over three trials with different random seeds.

\subsubsection{Dataset Statistics}
The following datasets are used in baseline comparisons:
\begin{itemize}
\item \textbf{RotationMNIST~(R-MNIST)} is a variant of MNIST~\cite{lecun1998mnist} where each of 36 tasks corresponds to images rotated by a multiple of 10 degrees. The ground truth labels remain the same. 
\item \textbf{PermutedMNIST~(P-MNIST)}~\cite{kirkpatrick2017overcoming} is another variant of MNIST where new tasks are created with a deterministic permutation of the pixels of MNIST. The total number of tasks we test is 100, and the ground truth labels remain the same. 
\item \textbf{Split-CIFAR100} is constructed by randomly splitting CIFAR100~\cite{krizhevsky2009learning} into 20 different 5-way classification problems.
\item \textbf{Split-miniImageNet}~\cite{chaudhry2019tiny} splits the mini-ImageNet dataset~\cite{vinyals2016matching}, a subset of ImageNet~\cite{deng2009imagenet} with 100 classes, into 20 disjoint tasks where each task contains 5 classes.
\end{itemize}
Table~\ref{tab:dataset_stat} depicts the statistics of all benchmark dataset above.
\begin{table*}[t!]
\caption{Benchmark Dataset Statistics}
\centering
\begin{tabular}{m{0.2\textwidth}|m{0.15\textwidth}|m{0.15\textwidth}|m{0.15\textwidth}|m{0.18\textwidth}}
\toprule
\textbf{Dataset} & \textbf{R-MNIST} & \textbf{P-MNIST} & \textbf{Split CIFAR100} & \textbf{Split-miniImageNet} \\
\hline
\hline
Num. of tasks & 36 & 100 & 20 & 20 \\ 
\hline
Input size & $1 \times 28 \times 28$ & $1 \times 28 \times 28$ & $3 \times 28 \times 28$ & $3 \times 84 \times 84$ \\ 
\hline
\# Classes/task & 10 & 10 & 5 & 5 \\ 
\hline
\# Training samples/task & 60,000 & 60,000 & 2,500 & 2,500 \\ 
\hline
\# Test samples/task & 10,000 & 10,000 & 500 & 500 \\
\bottomrule
\end{tabular}
\label{tab:dataset_stat}
\end{table*}

\subsection{Architecture Details}
\subsubsection{Single-hidden-layer network for \textsc{FlyNet}} 
For the experiments on R-MNIST and P-MNIST, this architecture is applied only to our \textsc{FlyNet}. In order to show the performance change according to the size of the hidden layer, we test the different sizes of hidden layers by multiplying the input size by the scaling factors $\{ 10d, 20d, 30d, 40d \}$. For example, $10d = \text{Input Size} \times 10$.

\subsubsection{Two-hidden-layer Fully Connected~(FC) network}
For the experiments on R-MNIST and P-MNIST, we use the modified experimental setup from~\cite{wortsman2020supermasks} to make other baselines have fewer learnable parameters. We use a two-hidden-layer FC network of 256 units per hidden layer~(FC256) for other baseline methods. For our \textsc{NeuroModNet}, we use a two-hidden-layer FC network with a set of different unit sizes per hidden layer, $\{ 256, 512, 1024, 2048, 4096 \}$, to show the trade-off between the number of training parameters and performance.

\subsubsection{Reduced ResNet-18}
For the experiments on Split-CIFAR100, we use a smaller version of ResNet-18 with two times fewer feature maps across all layers than the ResNet-18 used in~\cite{wortsman2020supermasks}. This reduced ResNet-18 is for other baseline models and, based on the architecture, we define the following scaling factor that is multiplied with the feature maps across all layers of the model for our \textsc{NeuroModNet}, $\{ 1d, 2d, 4d, 6d \}$, so that the model can have a different number of learnable parameters. For example, $2d$ has two times larger feature maps across all layers than $1d$.

\subsubsection{Standard ResNet-18}
For the experiments on Split-miniImageNet, we use a standard version of ResNet-18~\cite{he2016deep} for all models, including our \textsc{NeuroModNet} and other baselines. 

\subsection{Training Details and Hyperparameter Settings}
In training all the models within the baseline comparison, we follow the experimental setup from~\cite{wortsman2020supermasks}.

For the experiments on R-MNIST and P-MNIST, we train for 1000 batches of size 128 using the RMSProp optimizer~\cite{tieleman2012lecture} for all models. Specifically, for \textsc{PSP}~\cite{cheung2019superposition}, we utilize the rotational superposition, the best performing model from \textsc{PSP}, and the same hyperparameters from its official code. For \textsc{BatchE}~\cite{wen2020batchensemble}, we use the same implementation from~\cite{wortsman2020supermasks} as follows: using kaiming normal initialization with a learning rate of 0.01 (0.0001 for the first task when the weights are trained). For~\textsc{SupSup}, we use its official implementation with the learning rate of 0.0001 but without the module for inferencing task boundary, which allow it to achieve the best performance. For our CRWN, both \textsc{FlyNet} and \textsc{NeuroModNet} uses a learning rate of 0.01, which is very larger than that for other models. 

For the experiments on Split-CIFAR100, we train each model for 250 epochs per task. We use the Adam optimizer~\cite{kingma2014adam} with a batch size of 128. Baseline models use a learning rate of 0.001 (no warmup, cosine decay~\cite{loshchilov2016sgdr}). In particular, for \textsc{PSP}, we use the binary superposition PSP model from the official implementation~\cite{cheung2019superposition} so that it can be applied to a reduced ResNet-18. For \textsc{BatchE}, we use the \textsc{BatchE} with the same learning rates as the one used in the experiments on variants MNIST. For~\textsc{SupSup}, we utilize the official implementation from~\cite{wortsman2020supermasks}. For \textsc{NeuroModNet}, we use a learning rate of 0.1 to train. 

For the experiments on Split-miniImageNet, there are no experimental setups provided by the references of baseline models. Thus, we utilize the exact setup of the experiments on Split-CIFAR100 again so there is no difference in experimental setups between the Split-CIFAR100 and Split-miniImageNet.

\end{document}